\documentclass[conference]{IEEEtran}

\makeatletter
\def\ps@IEEEtitlepagestyle{%
  \def\@oddfoot{\mycopyrightnotice}%
  \def\@oddhead{\hbox{}\@IEEEheaderstyle\leftmark\hfil\thepage}\relax
  \def\@evenhead{\@IEEEheaderstyle\thepage\hfil\leftmark\hbox{}}\relax
  \def\@evenfoot{}%
}
\def\mycopyrightnotice{%
  \begin{minipage}{\textwidth}
  \scriptsize
  \copyright~2021 IEEE. Personal use of this material is permitted. Permission from IEEE must be obtained for all other uses, in any current or future media, including reprinting/republishing this material for advertising or promotional purposes, creating new collective works, for resale or redistribution to servers or lists, or reuse of any copyrighted component of this work in other works. 
  
  This work has been accepted at The Asia and South Pacific Design Automation Conference (ASP-DAC’22).
  \end{minipage}
}
\makeatother

\usepackage{geometry}
\ifCLASSINFOpdf
\else
\fi
\usepackage{cite}
\usepackage{amsmath,amssymb,amsfonts}
\usepackage{graphicx}
\usepackage{textcomp}
\usepackage{xcolor}

\usepackage{booktabs}
\usepackage[binary-units=true]{siunitx}
\usepackage{array,multirow}
\usepackage{hyperref}
\usepackage{amsmath}
\usepackage{enumitem}
\usepackage{algorithm}
\usepackage{graphicx}
\usepackage{float}

\usepackage[noend]{algpseudocode}

\newcommand{\figref}[1]{Figure~\ref{#1}}
\newcommand{\secref}[1]{Section~\ref{#1}}

\newcommand{\tabref}[1]{Table~\ref{#1}}

\newcommand{\Hsection}[1]{\vspace{0.5\baselineskip}\par\noindent\textit{#1}~\textbf{---}~}

\newcolumntype{L}[1]{>{\raggedright\let\newline\\\arraybackslash\hspace{0pt}}m{#1}}
\newcolumntype{C}[1]{>{\centering\let\newline\\\arraybackslash\hspace{0pt}}m{#1}}
\newcolumntype{R}[1]{>{\raggedleft\let\newline\\\arraybackslash\hspace{0pt}}m{#1}}

\begin{document}
%
\author{
\IEEEauthorblockN{Hongxiang Fan\IEEEauthorrefmark{1}, Martin Ferianc\IEEEauthorrefmark{2}, Zhiqiang Que\IEEEauthorrefmark{1}, He Li\IEEEauthorrefmark{6}, Shuanglong Liu\IEEEauthorrefmark{3}, Xinyu Niu\IEEEauthorrefmark{4}, Wayne Luk\IEEEauthorrefmark{1}}
\IEEEauthorblockA{\IEEEauthorrefmark{1}
Dept. of Computing, School of Engineering, Imperial College London, UK\\
\textit{\{h.fan17, z.que, w.luk\}@imperial.ac.uk}}
\IEEEauthorblockA{\IEEEauthorrefmark{2}
Dept. of Electronic and Electrical Engineering, University College London, UK, \textit{martin.ferianc.19@ucl.ac.uk}}
\IEEEauthorblockA{\IEEEauthorrefmark{6}
Dept. of Engineering, University of Cambridge, Cambridge, UK, \textit{he.li@ieee.org}} 
\IEEEauthorblockA{\IEEEauthorrefmark{3}
Hunan Normal University, Changsha, China, \textit{liu.shuanglong@hunnu.edu.cn}} 
\IEEEauthorblockA{\IEEEauthorrefmark{4}
Corerain Technologies Ltd., Shenzhen, China, \textit{xinyu.niu@corerain.com}} 
}
\title{Algorithm and Hardware Co-design  for Reconfigurable CNN Accelerator}

\maketitle

\begin{abstract}
Recent advances in algorithm-hardware co-design for deep neural networks (DNNs) have demonstrated their potential in automatically designing neural architectures and hardware designs.
Nevertheless, it is still a challenging optimization problem due to the expensive training cost and the time-consuming hardware implementation, which makes the exploration on the vast design space of neural architecture and hardware design intractable.
In this paper,
we demonstrate that our proposed approach is capable of locating designs on the Pareto frontier.
This capability is enabled by a novel three-phase co-design framework, with the following new features: 
(a) decoupling DNN training from the design space exploration of hardware architecture and neural architecture, 
(b) providing a hardware-friendly neural architecture space by considering hardware characteristics in constructing the search cells,
(c) adopting Gaussian process to predict accuracy, latency and power consumption to avoid time-consuming synthesis and place-and-route processes.
In comparison with the manually-designed ResNet101, InceptionV2 and MobileNetV2, we can achieve up to 5\% higher accuracy with up to 3$\times$ speed up on the ImageNet dataset.
Compared with other state-of-the-art co-design frameworks, our found network and hardware configuration can achieve
2\% ${\sim}$ 6\% higher accuracy, 2$\times {\sim}$ 26$\times$ smaller latency and 8.5$\times$ higher energy efficiency.
\end{abstract}


%
\IEEEpeerreviewmaketitle

\section{Introduction}

The success of deep learning, and especially neural networks (NNs), has attracted enormous research and industrial interests in applying NNs in real-life scenarios such as in autonomous driving~\cite{grigorescu2020survey}.
However, the heavy computational and memory demand of running NNs imposes a large overhead on their hardware performance, in particular while considering resource-constrained platforms~\cite{fan2018real}.
Currently, there are two research directions that focus on improving the hardware performance of deployed NNs.
First, algorithm-level design of efficient NNs through neural architecture search (NAS)~\cite{zoph2016neural}, which automatically designs NN architectures with high accuracy and low computational complexity for different scenarios~\cite{wu2019fbnet}. 
Second, hardware-level efforts to design highly-optimized and specialized hardware accelerators for NNs~\cite{fan2019f,liu2021toward,yu2020collaborative}.
However, most of the time, the algorithm-level optimization and the hardware-level design are not considered jointly, which can lead to sub-optimal solutions in terms of both the resultant algorithmic or the hardware performance.
For example, the authors in~\cite{fan2018real} demonstrate that the hardware architecture designed for the regular convolution is not suitable for depthwise convolution commonly used in NAS~\cite{wu2019fbnet}.

To address the aforementioned sub-optimality, there is a growing demand for a method that can perform NAS to design accurate NNs and at the same time, co-develop hardware designs customized for the NN found.
To meet this demand, reconfigurable hardware, such as field-programmable gate arrays (FPGAs), represents an ideal platform to implement algorithm-hardware co-design. Given its reconfigurability, FPGA can be utilized to provide highly-optimized hardware, customized for different NNs found by NAS.
{
Previous work has attempted to apply evolutionary algorithm~\cite{linneural, colangelo2019artificial, colangelo2019evolutionary}, reinforcement learning~\cite{jiang2019hardware, abdelfattah2020best} and differentiable NAS~\cite{li2020edd, fan2020optimizing} on algorithm-hardware co-design for NNs on FPGA.
However, these approaches iteratively perform the network training and design space exploration for multiple times, making the process time-consuming.
Also, the characteristics of the accelerator are only considered during NAS in their work.
To address these issues, our contributions include:
}

\begin{itemize}[leftmargin=*]
  \item {A novel three-phase co-design framework, which decouples network training from design space exploration of both hardware design and neural architecture to avoid iterative time-consuming optimization.
  A hardware-friendly neural architecture space is also proposed by considering the characteristics of the underlying hardware to construct search cells before the neural architecture searching (\secref{sec:nas});}
  \item An accurate and efficient cross-entropy loss, latency and energy consumption model based on Gaussian process regression, together with a genetic algorithm, which enable fast design space exploration within few minutes (\secref{sec:design_space});
  \item A demonstration of the effectiveness of the proposed method on the ImageNet dataset. The network found and its custom hardware design lie in the Pareto frontier, and can achieve better accuracy, energy efficiency and latency in comparison to other state-of-the-art co-design methods
  (\secref{sec:exp}).
\end{itemize}

\begin{figure*}\centering
\includegraphics[width=0.9\textwidth]{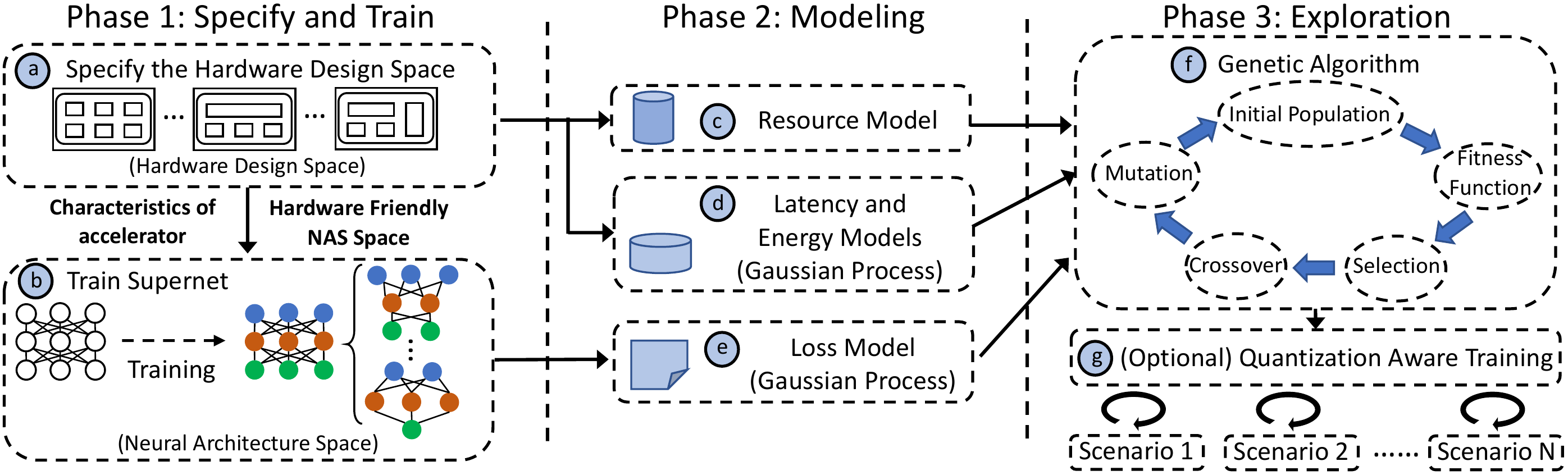}
\caption{The overview of the proposed framework.}\label{fig:overview_framework}
\end{figure*}

\section{Background}\label{sec:background}

\subsection{Algorithm and Hardware Co-design}\label{subsec:codesign}
The joint design of NNs and hardware has been recently an activate research area~\cite{hao2019fpga}.
Based on an evolutionary algorithm (EA),
Lin \textit{et al.}~\cite{linneural} propose a two-stage method for algorithm-hardware co-design.
{Although their work claims $1.3\times$ speedup and $1.6\times$ energy savings, their results are based entirely on simulations, and the performance is estimated without running on a real hardware.
At the same time, EA has been adopted in other NAS methods~\cite{colangelo2019artificial, colangelo2019evolutionary}.
However, training cost is expensive and the generated NNs lack in accuracy.
}

Reinforcement learning (RL) is another approach used for the algorithm-hardware co-design~\cite{jiang2019hardware}.
Nevertheless, a common drawback in these RL-based algorithm-hardware co-design approaches is that they demand significant number of GPU hours for search of both algorithm and hardware-defining parameters,
which is unfeasible for real-life applications.
To reduce the search cost,
differentiable NAS (DNA) has been used in algorithm and hardware co-design~\cite{li2020edd}.
However,
it has been demonstrated in~\cite{li2020random} that the NNs found by DNA can only achieve similar accuracy to the NNs generated by random search.

{
The once-for-all (\textit{OFA}) proposed in~\cite{cai2019once} provides another paradigm for NAS.
The progressive shrinking algorithm has been demonstrated to be effective in training the supernet.
However, their work only optimizes neural architectures without searching the optimal hardware architectures.
Also the neural architecture space in their paper does not consider the characteristics of underlying hardware design, which leads to sub-optimal hardware performance.
Compared with their work, we are able to achieve a higher accuracy and hardware performance as demonstrated in~\secref{sec:exp}.
Although~\cite{lin2021naas} tries to search the accelerator architecture,
they only focus on processing engine (PE) connectivity and compiler mappings for an ASIC design.
}

\subsection{Gaussian Process}\label{sec:background_gp}
Gaussian process (GP) is a model built around Bayesian probabilistic theory which can embody prior knowledge into the predictive model and can be used for regression of real valued non-linear targets~\cite{rasmussen2010gaussian}. A GP is specified by a mean function and a covariance function-kernel. A common choice for kernels includes polynomial, Gaussian or Matérn kernels~\cite{rasmussen2010gaussian}. The mean function represents the supposed average of the estimated data. The kernel computes correlations between inputs and it encapsulates the structure of the hypothesised function. 
GP allows fast estimation, which is especially useful in design space exploration~\cite{ferianc2020improving}.
However,~\cite{ferianc2020improving} only explored latency estimation for a single layer, while this paper adopts GP to estimate the loss, latency and energy consumption of the whole NN.

\section{Algorithm-Hardware Co-design}\label{sec:nas}

The problem of algorithm-hardware co-design can be defined as follows:
\begin{equation}\label{eq:form_problem}
\begin{aligned} 
    \min _{\alpha \in \mathcal{A}, \beta \in \mathcal{B}} \min _{\boldsymbol{w}_{\alpha}} \mathcal{L}\left(\boldsymbol{w}_{\alpha}, \alpha, \beta \right)
\end{aligned}
\end{equation}
The $\mathcal{A}$ denotes the NN architecture space and $\mathcal{B}$ represents the hardware design space.
To minimize the loss $\mathcal{L}$,
we aim to find the optimal hardware configuration $\beta \in \mathcal{B}$ and NN architecture $\alpha \in \mathcal{A}$ with the associated weights $\boldsymbol{w}_{\alpha}$.

In this paper,
we decouple the training of weights $\boldsymbol{w}_{\alpha}$ and the optimization of $\alpha$ and $\beta$ into two separate steps.
First, we train a \textit{supernet}, encompassing all our NN architecture options, with respect to the weights $\boldsymbol{w}_{\alpha}$ using the following objective function based on a cross-entropy (CE) loss:
\begin{equation}\label{eq:fst_problem}
\begin{aligned}
    \min _{\boldsymbol{w}_{\alpha}} \sum_{\alpha \in \mathcal{A}} 
    CE\left(\boldsymbol{w}_{\alpha}, \alpha\right).
\end{aligned}
\end{equation}
During this process, we randomly sample sub-NNs from the supernet and independently train each sampled network to minimize the overall loss.
Then,
once the training is finished,
we perform the optimization with respect to the $\alpha$ and $\beta$ using the overall objective function $\mathcal{L}$ containing both the CE loss and hardware costs as follows:
\begin{equation}\label{eq:snd_problem}
\begin{aligned}
    \min _{\alpha \in \mathcal{A}, \beta \in \mathcal{B}} \mathcal{L}\left( \alpha, \beta \right).
\end{aligned}
\end{equation}

Aiming at solving~\eqref{eq:fst_problem} and~\eqref{eq:snd_problem}, a novel algorithm-hardware co-design framework is proposed,
which is illustrated in \figref{fig:overview_framework}.
To make sure the framework is applicable to any reconfigurable hardware system,
we generalize it into three phases: \textit{1)} Specify and Train, \textit{2)} Modeling and \textit{3)} Exploration.
Note that the first and second phases are only required once, while the Exploration is briefly performed given a specific deployment scenarios, which makes our framework efficient.

\Hsection{\bf {Phase 1: Specify and Train}}To define the hardware design space for exploration,
it first requires the users to specify a reconfigurable hardware system to accelerate NNs.
Then, the neural architecture search space is built based on the supported operations provided by the underlying hardware system. The neural architecture search space often considers different algorithmic configurations, ordering and connections between operations inside the NNs~\cite{zoph2016neural}.
Note that, our framework does not apply any restrictions on the neural architecture space, and it can be changed accordingly for different reconfigurable hardware designs.
Therefore, our framework is general enough to cover any reconfigurable hardware, and has potential to gain higher accuracy and hardware performance.

During the training,
in order to efficiently solve~\eqref{eq:fst_problem},
we use the progressive shrinking algorithm~\cite{cai2019once} to train all the sub-networks within the supernet by random sampling of candidate NNs.
All the sub-networks share the same set of parameters within the supernet.
Once the training is finished,
we can quickly sample a sub-network from the supernet without extra effort.
All these sub-networks form the final neural architecture space, which enables exploration at the algorithm-level in later phase.

\Hsection{\bf {Phase 2: Modeling}}In the second phase,
we model different metrics: CE loss ($CE$), latency, energy and resource consumption, to enable fast exploration in the last phase.

For loss, latency and energy models, 
we adopt the GP regression for fast estimation.
The training data used for GP regression is obtained by randomly sampling a small number of sub-networks from the supernet.
These sampled NNs are then evaluated on the dataset to get the CE loss, and run on our reconfigurable hardware with different configurations to obtain their latency and energy consumption.
For the resource model,
we propose to use a simple analytic formulation to estimate the DSP and memory resource consumption.

\Hsection{\bf {Phase 3: Exploration}}In the last phase,
as the regression models for loss, latency, energy and resources are available,
Genetic algorithm (GA) is adopted for fast design space exploration in both neural architecture and hardware design search spaces.
Our GA contains five operations, i.e., population initialization, fitness function evaluation, selection, crossover and mutation.
The population is initialized by the randomly generated neural architectures and hardware designs.
The loss (fitness) function is defined as follows:
\begin{equation}
    \begin{footnotesize}
    \begin{aligned} 
        \mathcal{L}= \eta \times CE + \mu \times Latency + \lambda \times Energy + Res_{PT}.
    \label{eq:score}
    \end{aligned}
    \end{footnotesize}
\end{equation}

The $\eta$, $\mu$ and $\lambda$ are user-defined hyper-parameters which denote the importance for the CE loss, latency and energy consumption.
The $CE$, $Latency$ and $Energy$ are the regression results of GP-based CE loss, latency and energy models.
Different values for $\eta$, $\mu$ and $\lambda$ may lead to different results and this is explored in Section~\ref{sec:exp}.
The $Res_{PT}$ term is defined as follows:
\begin{equation}
  \begin{footnotesize}
    \begin{aligned} 
    Res_{PT} = 
    \begin{cases}
     0,\enskip  DSP_{used} \leq DSP_{avl}, MEM_{used} \leq MEM_{avl}  \\
     \gamma ,\enskip DSP_{used} > DSP_{avl}, MEM_{used} > MEM_{avl}  \\
    \end{cases}
    \label{eq:pm_time_input_data}
    \end{aligned}
  \end{footnotesize}
\end{equation}
where the $DSP_{avl}$ and $MEM_{avl}$ represent the available DSPs and memory resources on the target hardware platform and $DSP_{used}$ and $MEM_{used}$ denote DSP and memory consumption provided by the resource model.
The $\gamma$ denotes the penalty added to the loss function when the hardware resource consumption exceeds the budget.
In our very last step, if the underlying hardware supports different precision other than the one used for training the supernet,
the quantization aware finetuning~\cite{jacob2018quantization} will be enabled to tailor the resultant NNs to the hardware system.
The process of GA is illustrated in~\figref{fig:overview_framework}.

\section{Design Space and Modelling}\label{sec:design_space}

\subsection{Design Space}\label{subsec:design_space}
The design space is composed of two parts: hardware design space and neural architecture space.
\subsubsection{Hardware Design Space}
This paper adopts an example design which uses a single configurable processing unit to process different layers.
{
Although there are other designs, such as the streaming design~\cite{li2020edd, hao2019fpga} with layer-wise
reconfigurability,
they usually require a large amount of on-chip memory to cache all the intermediate results, which restricts the model size of the NN and limits the neural architecture space. In this paper, we adopted the single processing engine design, such that our search space encompasses larger CNNs.
Note that, although we use the single engine design, our framework is general enough to be applied to any reconfigurable design such as the streaming design by changing the hardware design space.
}

\begin{figure}[t]\centering
\includegraphics[width=0.36\textwidth]{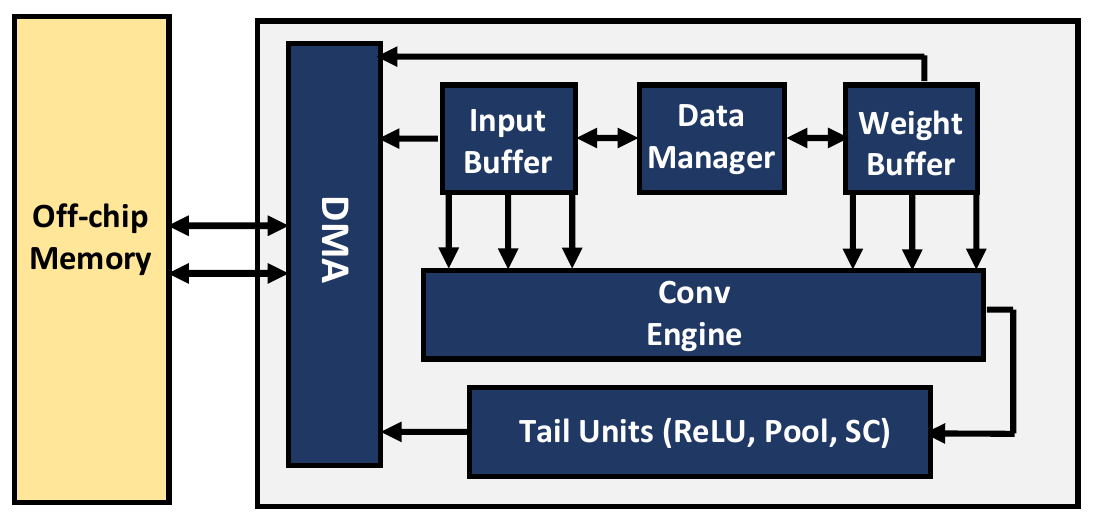}
\vspace{-1.5mm}
\caption{{Overview of the FPGA-based accelerator.}}\label{fig:hw}
\vspace{-4.5mm}
\end{figure}

The adopted reconfigurable design is illustrated in~\figref{fig:hw}.
The accelerator consists of an input buffer, a weight buffer, a convolutional ({\textit {Conv}}) engine and other functional modules including Shortcut (\textit{SC})~\cite{he2016deep}, Pooling (\textit{Pool}) and Rectified linear unit (\textit{ReLU}) activation.
The computation of the NN is performed sequentially, layer-by-layer, 
and only one layer is processed in the {\textit {Conv}} engine at a time.
This computational pattern allows the accelerator to support NNs even with a large number of layers because only one layer's input data and weights need to be cached in the on-chip memory.
To achieve higher hardware performance, the accelerator is designed to support
8-bit integer operations.

The {\textit {Conv}} engine supports three types of configurable parallelism: filter parallelism ($PF$), channel parallelism ($PC$) and vector parallelism ($PV$).
Different types of convolutions require different combinations of $PF$, $PC$ and $PV$ to achieve an optimal performance.
For instance, a convolution with small number of channels can achieve lower latency with the combination of low $PC$ and high $PF$ and $PV$ values, since there is no available concurrency in the channel dimension.
Our hardware design space is represented by memory size $MEM$, bandwidth $BW$ and different parallelism levels including $PF$, $PC$ and $PV$.
The domain for both $PF$ and $PC$ is $\{8, 16, 32, 64, 128\}$ and $PV$ can be chosen from $\{4, 8, 16\}$.
$MEM$ depends on the available memory resources on the FPGA board and $BW$ is selected from $\{32, 64, 128, 256\}$bits.
Thus, there are totally $5 \times 5 \times 3 \times 4$ potential different configurations for the hardware design.

\subsubsection{Neural Architecture Space}
\begin{figure}[t]\centering
\includegraphics[width=0.48\textwidth]{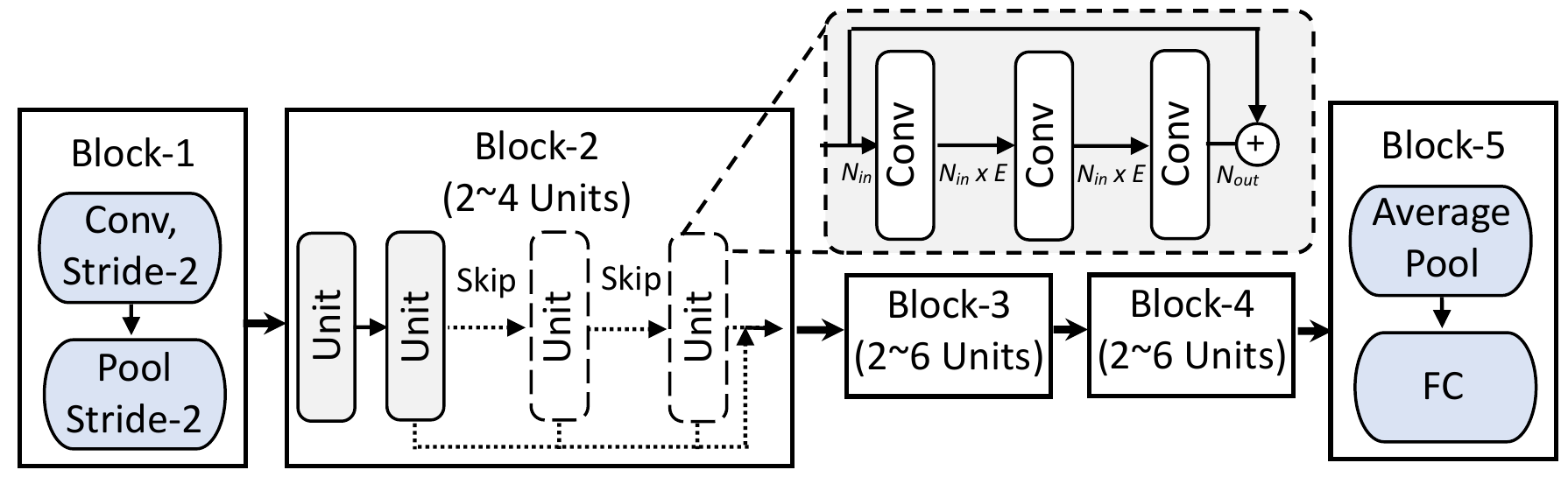}
\vspace{-1.5mm}
\caption{{The search space of neural architectures.}}\label{fig:nn_space}
\vspace{-4.5mm}
\end{figure}

The example architecture space is illustrated in~\figref{fig:nn_space}.
{
In this paper,
we argue that the design of neural architecture search should consider the underlying hardware design before the NAS optimization.
By analyzing the characteristics of the selected hardware architecture,
we found that it is efficient in performing the regular convolution with residual addition~\cite{he2016deep}.
Also,~\cite{bello2021revisiting} demonstrate that the basic building block of \textit{ResNet} is still one of the most effective architectures with the proper scaling strategies.
Therefore,
our core neural architecture search space follows the backbone of {\textit{ResNet-50}} which is composed of four residual blocks with gradually reduced feature map size and increased channel sizes.}
In each block, we search for the number of units ranging from 2 to $U_{i}$, where $U_{i}$ denotes the maximal number of units in $i^\textsuperscript{th}$ block.
In each cell, we search for the expansion ratio ($E$) chosen from \{$0.5$, $0.75$, $1.0$\}.
As there are totally $16$ cells in our neural architecture space,
the total number of combinations is $3^{16}$. Together with the 300 different hardware configuration options, there are more than $12 \times 10^{9}$ different combinations in our co-design space.

\subsection{Loss, Latency, Energy and Resource Models}\label{subsec:models}
\subsubsection{Loss Model}
Evaluating $CE$ in Equation~\ref{eq:fst_problem} for all 12 billion configurations on a large dataset such as ImageNet~\cite{deng2009imagenet} is time-consuming.
To enable fast evaluation, we adopt GP regression to estimate the $CE$ for all sub-networks.
To represent the neural architecture,
we encode the neural architecture space, which contains 16 searchable cells, into a 16-dimension vector
with each dimension representing the expansion ratio used in that cell.
The expansion ratio is 0, if a cell is skipped.
We construct a training dataset by randomly sampling and evaluating a certain number of sub-networks.
Based on the encoded input vector and evaluated $CE$,
we perform regression using the GP model with a Matérn covariance kernel with a constant mean function.

\subsubsection{Latency and Energy Models}
Measuring the hardware performance of all sub-networks for the FPGA-based design for different design parameters is time-consuming because of synthesis and place and route processes that are needed for the real hardware implementation.
We again use GP regression model to estimate the latency and energy consumption.
To represent the NN together with the hardware configuration, we encode it into a 19-dimensional vector with the first 16 dimensions representing the neural architecture and the last 3 dimensions being $PF$, $PC$ and $PV$.

\subsubsection{Resource Model}
As DSPs and memory are the limiting resource for FPGA-based CNN accelerator~\cite{liu2018optimizing}, we primarily consider DSP and memory consumption in this paper. 
The DSP consumption can be described as: $DSP_{used} = (PC \times PF \times PV)/2$,
which is dominated by the parallelism level used in the {\textit{Conv}} engine.

The memory resources are mainly consumed by the input and weight buffers.
As the input buffer needs to cache all the input feature maps in the current $i$\textsuperscript{th} layer, its usage can be represented as: $MEM_{in} = \max\limits_{i = 1 ,\ldots, l}(N_{c}^{i} \times H^{i} \times W^{i}) \times DW$, where $N_{c}^{i}$, $H^{i}$ and $W^{i}$ denote the number of channels, height and width of the input feature map, $DW$ is the data width and $l$ is the total depth of the net.
As for the weight buffer,
because weights are shared along $PV$ parallelism,
it only needs to cache the current $PF$ filters,
so the memory consumption can be formulated as: $MEM_{weight} = \max\limits_{i = 1,\ldots, l}(N_{c}^{i} \times PF \times K^{i} \times K^{i}) \times DW $, where $K^{i}$ is the kernel size of the $i$\textsuperscript{th} layer.
Due to the use of ping-pong buffer technique, the total memory consumption is: $MEM_{used} =  2 \times (MEM_{in} + MEM_{weight})$.

\section{Experiments}\label{sec:exp}
The PyTorch and GPyTorch libraries are used for the implementation of the supernet training and the GP models respectively.
ImageNet~\cite{deng2009imagenet} dataset contains over 10,000,000 labeled images of 1000 object categories for classification.
The hardware design used in all experiments is implemented on an Intel Arria 10 SX660 FPGA platform using Verilog. 1GB DDR4 SDRAM is installed on the platform as the off-chip memory. Quartus 17 Prime Pro was used for synthesis and implementation.
An Intel Xeon E5-2680 v2 CPU was used as the host processor. {We train the supernet on a GPU cluster with six NVIDIA GTX 1080 Ti GPUs for $4$ days.} A power meter is plugged in to measure the runtime power
performance.

\subsection{Accuracy of Gaussian Process-based Model}
To train our GP-based loss model, 2000 sub-networks were sampled and evaluated on ImageNet~\cite{deng2009imagenet}.
We used 1500 samples for training and 500 samples for evaluation.
The model was trained for 50 iterations using an Adam optimizer.
The result is shown in~\tabref{tb:regression_model}.
The mean absolute error (MAE) is only $0.01005$, which demonstrates the GP-based loss model is sufficiently accurate for the modeling.

\begin{table}[H]
\centering
\caption{Results of Gaussian process-based models.}
\label{tb:regression_model}
\scalebox{0.95}{
\setlength\tabcolsep{6pt} 
\begin{tabular}{c|c|c}
\toprule
& {\bf  Kernel Function} & {\bf Mean Absolute Error}\\
\midrule
{\bf Loss Model}& Matérn ($3 / 2$)& 0.01005\\
\midrule
{\bf Latency Model}& Matérn ($5 / 2$) & 0.06521ms\\
\midrule
{\bf Energy Model}& Matérn ($5 / 2$) & 0.01804W\\
\bottomrule
\end{tabular}}
\vspace{-1.5mm}
\end{table}

Similarly, 4600 random samples with different network configurations and hardware designs were collected for latency and energy modeling.
We used 3000 and 1600 samples for training and evaluation respectively.
The training was again performed with respect to 50 iterations and an Adam optimizer.
As shown in~\tabref{tb:regression_model},
the MAE of our GP-based latency and energy models is only $0.06521$ms and $0.01804$W.
Therefore, the proposed GP-based latency and energy models can be used as an accurate estimator for the latency and energy consumption.

\subsection{Effectiveness of Design Space Exploration}
\begin{figure}[t]\centering
\includegraphics[width=0.35\textwidth]{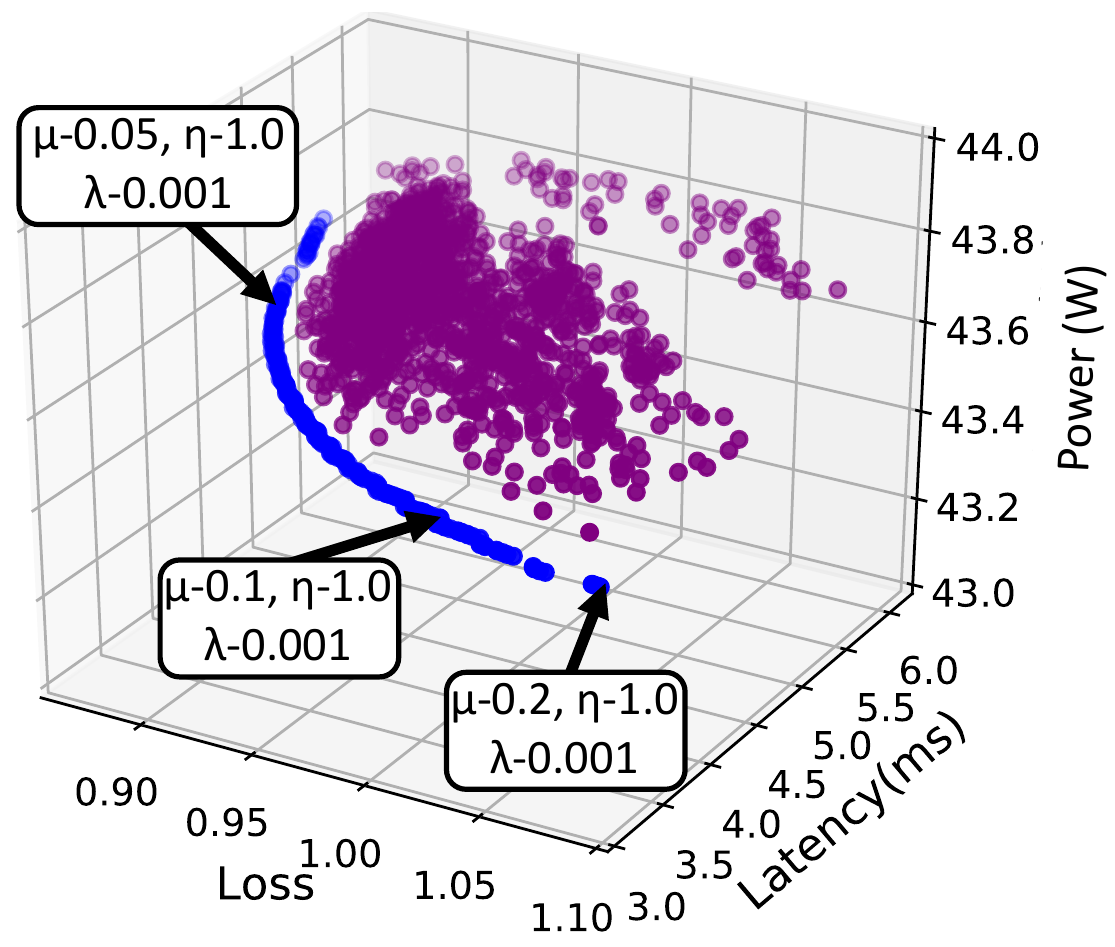}
\vspace{-2mm}
\caption{The performance of various NAS-generated NNs on different candidate hardware design. Pareto-optimal is denoted by blue points.}\label{fig:pt_space}
\vspace{-5mm}
\end{figure}

{For reference and demonstration,
we iterated through and evaluated all samples in the co-design space to get the reference Pareto frontier.}
The Pareto-optimal points, which are better in either loss or latency or energy with respect to any other point, form a Pareto frontier, which is drawn as blue points in~\figref{fig:pt_space}.
Because the whole design space is too large to show in the Figure.
we randomly drew 2000 non-Pareto-optimal samples as purple points to visualize the rest of the design space.

\begin{figure}[b]\centering
\vspace{-4mm}
\includegraphics[width=0.49\textwidth]{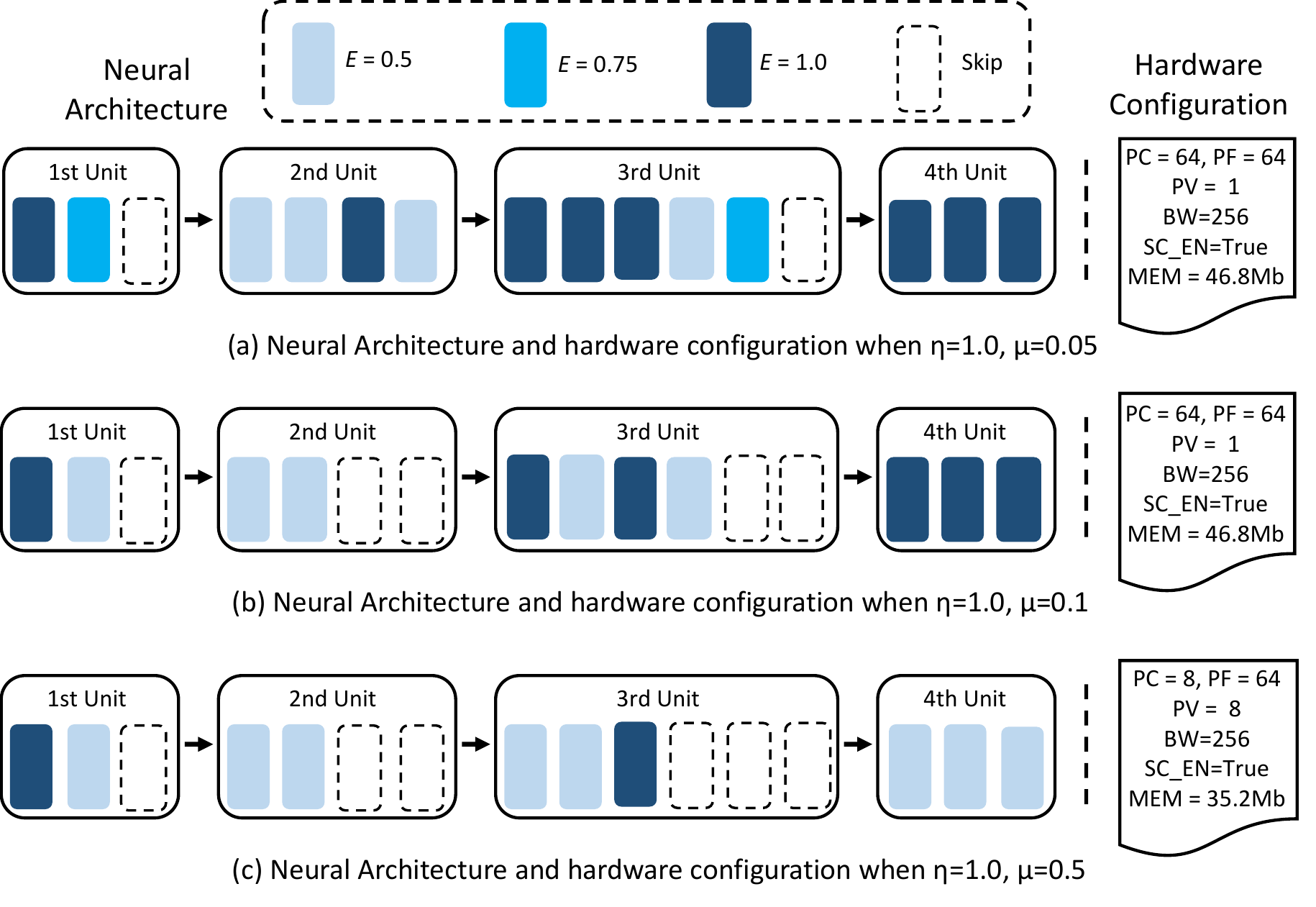}
\vspace*{-0.75cm}
\caption{Neural architecture and hardware configuration of NNs found.}\label{fig:found_nn}
\end{figure}

Then, to demonstrate the effectiveness of our framework, we used GA to perform design space exploration, and check whether these found configurations match the reference Pareto frontier.
{
The time cost for the proposed GP-based models and GA to find one optimized design is only $0.1$ GPU hour, which demonstrates the efficiency of our framework.
In contrast, other approaches~\cite{jiang2019hardware, dong2021hao} require tens to hundreds of GPU hours in searching}.
As mentioned in~\secref{sec:nas},
the user-defined hyper-parameters $\eta$, $\mu$ and $\lambda$ specified in the GA represent the importance of accuracy, latency and energy consumption respectively,
we chose three sets of $\eta$, $\mu$ and $\lambda$: $\{1.0, 0.2, 0.001\}$, $\{1.0, 0.1, 0.001\}$ and $\{1.0, 0.05, 0.001\}$, to demonstrate how the GA is able to find different Pareto-optimal designs according to users' requirements.
The resultant designs found by the GA are highlighted by black arrows in~\figref{fig:pt_space}, which all lay on the reference Pareto frontier.
Their NN architectures and hardware configurations are illustrated in~\figref{fig:found_nn}.
Therefore, our framework can effectively identify the Pareto-optimal designs in the vast algorithm-hardware co-design space.

We also evaluated the resultant networks on different hardware platforms including Intel Xeon Silver 4110 CPU and NVIDIA GTX 1080 Ti GPU.
The results are presented in~\tabref{tb:rnas_cpu_gpu}. 
{TensorRT and CuDNN $8.11$ libraries were used for GPU implementation, and the MKLDNN
was used to optimize the performance of the CPU implementation.} 
The batch size was set to one for a fair comparison.
Compared with GPU and CPU implementations, the networks found for the reconfigurable FPGA-based accelerator can achieve approximately $2\times$ and $6\times$ reduction in latency and up to $5\times$ and $19\times$ higher energy efficiency.

\begin{table}
\centering
\caption{Accuracy, latency and energy efficiency on ImageNet.}
\label{tb:rnas_cpu_gpu}
\scalebox{0.79}{
\setlength\tabcolsep{6pt} 
\begin{tabular}{c|c|c|c|c|c|c|c}
\toprule
&\multicolumn{2}{c|}{CPU}&\multicolumn{2}{c|}{GPU}&\multicolumn{2}{c|}{FPGA} &\multirow{3}{*}{\textbf{Acc}}\\
\cmidrule{2-7}
 & {Lat.} & {Enrg. Eff.} & {Lat.} & {Enrg. Eff.} & {Lat.} & {Enrg. Eff.} & \\
  & {(ms)} & {(FPS/W)} & {(ms)} & {(FPS/W)} & {(ms)} & {(FPS/W)} & \\
\midrule
{$\eta(1.0)\mu(0.05)$} & \multirow{2}{*}{$26.08$} & \multirow{2}{*}{0.28} & \multirow{2}{*}{$7.40$} & \multirow{2}{*}{0.94} & \multirow{2}{*}{4.52} & \multirow{2}{*}{5.07} & \multirow{2}{*}{\textbf{77.63}\%}\\
{$\lambda(0.001)$} &  &  &  &  &  &  & \\
\midrule
{$\eta(1.0)\mu(0.1)$} & \multirow{2}{*}{$24.06$} & \multirow{2}{*}{0.30} & \multirow{2}{*}{$6.57$} & \multirow{2}{*}{1.06} & \multirow{2}{*}{3.66} & \multirow{2}{*}{6.27} & \multirow{2}{*}{{76.30}\%}\\
{$\lambda(0.001)$} &  &  &  &  &  &  & \\
\midrule
{$\eta(1.0)\mu(0.2)$} & \multirow{2}{*}{$19.18$} & \multirow{2}{*}{0.38} & \multirow{2}{*}{$5.03$} & \multirow{2}{*}{1.38} & \multirow{2}{*}{\textbf{3.14}} & \multirow{2}{*}{\textbf{7.32}} & \multirow{2}{*}{{74.91}\%}\\
{$\lambda(0.001)$} &  &  &  &  &  &  & \\
\bottomrule
\end{tabular}}
\vspace{-3.5mm}
\end{table}

\subsection{Comparison with Manually Designed Networks}
\begin{figure}[t]\centering
\includegraphics[width=0.46\textwidth]{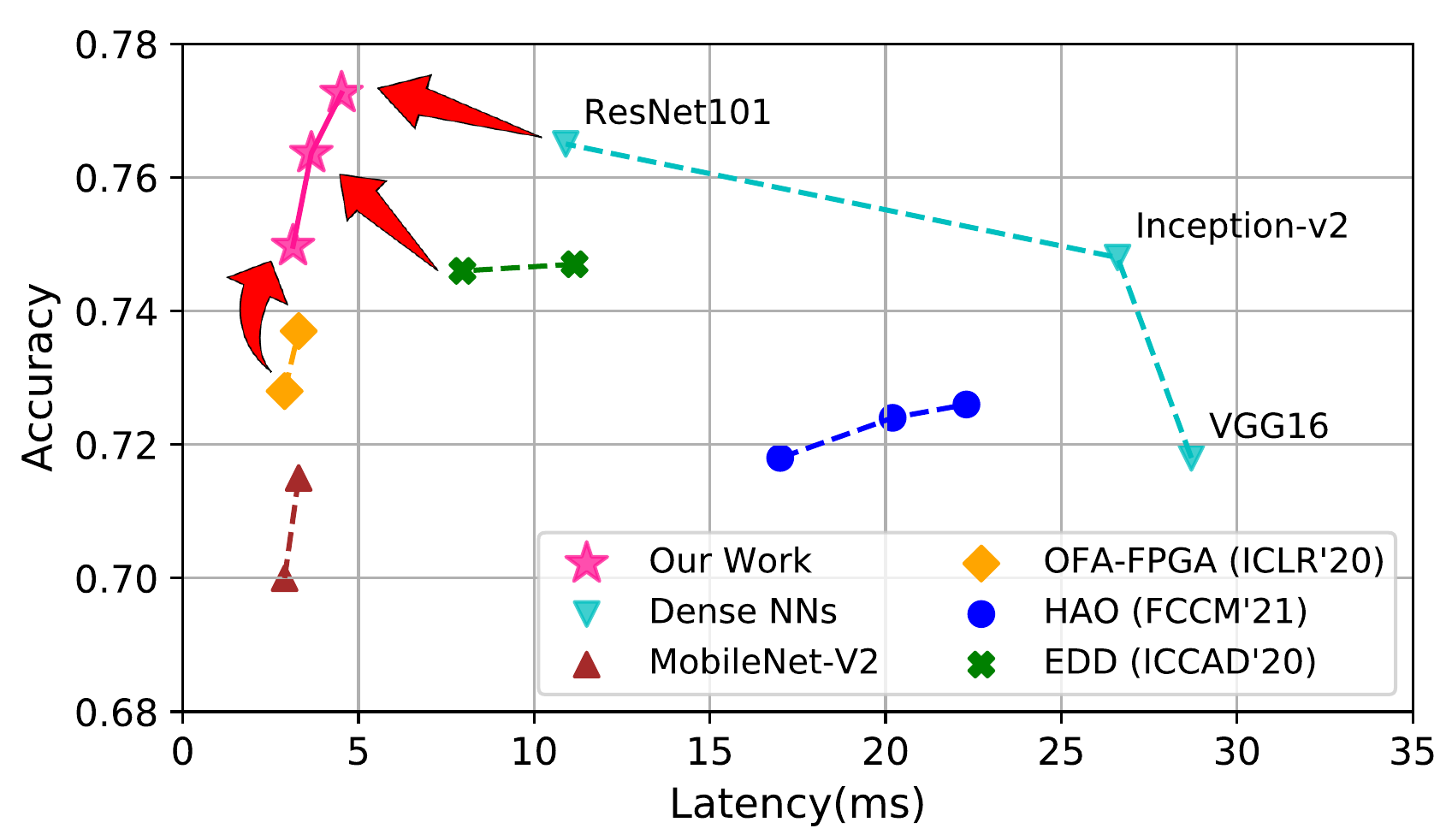}
\vspace*{-0.2cm}
\caption{{Comparison of accuracy and latency among our work, the manually-designed neural networks and other algorithm-hardware co-design methods.}}\label{fig:nns}
\end{figure}

To demonstrate that the auto-generated NN architectures can outperform manually-designed networks in terms of accuracy, latency, energy and model size on our FPGA accelerator,
we evaluated several commonly benchmarked NNs including \textit{ResNet-101}~\cite{he2016deep}, 
\textit{VGG-16}~\cite{simonyan2014very} and 
\textit{Inception-v2}~\cite{szegedy2016rethinking} 
on the ImageNet.
The hardware configurations with respect to these networks were manually optimized.
The results are shown in~\figref{fig:nns}.
The network found with highest accuracy ($\eta=1.0$, $\mu=0.05$, $\lambda=0.001$) is nearly $1\%$ more accurate and $3 \times$ faster than \textit{ResNet-101}.
Compared with \textit{VGG-16}, the network found can achieve nearly $5\%$ higher accuracy while reducing the latency by nearly $10\times$.
{We also compared our work with the \textit{MobileNetV2}~\cite{sandler2018mobilenetv2} implemented in~\cite{cai2019once}.
Our design achieves a similar latency while improving the accuracy by nearly $4$\%.}

\subsection{Comparison with Existing Co-Design Work}
{We compared our proposed approach with four other state-of-the-art co-design methods, including \textit{Co-Explore}~\cite{jiang2019hardware}, \textit{EDD}~\cite{li2020edd}, \textit{HAO}~\cite{dong2021hao}, and \textit{OFA}~\cite{cai2019once}.
Although there are other co-design works, they suffer from low accuracy~\cite{colangelo2019artificial, colangelo2019evolutionary, hao2019fpga, jiang2019accuracy_vs_eff} or only evaluated on a small dataset~\cite{abdelfattah2020best, chen2020you, fan2020optimizing}. 
Therefore, we did not include them in our comparison.
The results are shown in~\figref{fig:nns}.
\tabref{tb:comp_existing} summarizes their underlying hardware platforms and implementation details.
Compared with the network generated by~\cite{jiang2019hardware},
our network found can achieve $6$\% higher accuracy, more than
$26\times$ speed up and nearly $8\times$ higher energy efficiency.
We can also achieve nearly $4$\% higher accuracy than \textit{HAO}~\cite{dong2021hao} with better hardware performance even with the latency being normalized by the DSP consumption.
In comparison with \textit{OFA} that consumes nearly twice more DSPs,
we achieve a similar latency with $2.7$\% higher accuracy.
}

\begin{table}[t]
\centering
\caption{{Details of hardware implementations.}}
\label{tb:comp_existing}
\scalebox{0.82}{
\setlength\tabcolsep{6pt} 
\begin{tabular}{c|c|c|c|c|c}
\toprule
& \multirow{2}{*}{\bf Platform}& {\bf Number}&  {\bf Latency}  & \multirow{2}{*}{\bf Accuracy} & {\bf Energy Eff.} \\
&& {\bf of DSPs}& (ms)  &&(GOPS/W)\\
\midrule
{\textit{Co-Explore}~\cite{jiang2019hardware}} & Xilinx XC7Z015 & 150& 95.24  & 70.24\% & 0.74\\
\midrule
{\textit{EDD}~\cite{li2020edd}}& Xilinx ZCU102& 2520& 7.96  & 74.60\%&-\\
\midrule
{\textit{HAO}~\cite{dong2021hao}} &Xilinx ZU3EG & 360& 22.27  & 72.68\%&-\\
\midrule
{\textit{OFA}~\cite{cai2019once}}& Xilinx  ZU9EG& 2520& 3.30  & 73.60\%&-\\
\midrule
{Our Work}& Intel GX1150 & 1345 & 3.66& 76.30\%& 6.27\\
\bottomrule
\end{tabular}}
\vspace{-4mm}
\end{table}

\section{Conclusion}
This paper proposes a novel algorithm-hardware co-design framework for reconfigurable NN accelerators.
To reduce the search cost,
we adopt genetic algorithm and Gaussian process regression, which enables fast design space exploration within few minutes.
The network and hardware configuration generated by the proposed framework on our reconfigurable CNN accelerator can
achieve 1\% to 5\% higher accuracy while reducing the latency by 2$\times$ to 10$\times$ on the ImageNet dataset, in comparison with manually-designed NNs on the same hardware.
Compared with the other state-of-the-art algorithm-hardware co-design approaches,
our found NNs achieve better accuracy, energy efficiency, latency and search cost.
Future work includes expanding the search space with more choices of operations, integrating optimization for recurrent neural networks into the current optimization step and supporting end-to-end automation.

\section*{Acknowledgement}{
The support of the United Kingdom EPSRC (No. EP/L016796/1, EP/N031768/1, EP/P010040/1, EP/V028251/1 and EP/S030069/1), the National Natural Science Foundation of China (No. 62001165), Hunan Provincial Natural Science Foundation of China (No. 2021JJ40357), Changsha Municipal Natural Science Foundation (No. kq2014079), Corerain, Maxeler, Intel and Xilinx is gratefully acknowledged.
}

\bibliographystyle{ieeetr}
\bibliography{references.bib}

\end{document}